# AN AUTOMATIC METHOD OF FINDING TOPIC BOUNDARIES

**Jeffrey C. Reynar***
Department of Computer and Information Science
University of Pennsylvania
Philadelphia, Pennsylvania, USA
jcreynar@unagi.cis.upenn.edu

## Abstract

This article outlines a new method of locating discourse boundaries based on lexical cohesion and a graphical technique called dotplotting. The application of dotplotting to discourse segmentation can be performed either manually, by examining a graph, or automatically, using an optimization algorithm. The results of two experiments involving automatically locating boundaries between a series of concatenated documents are presented. Areas of application and future directions for this work are also outlined.

## Introduction

In general, texts are "about" some topic. That is, the sentences which compose a document contribute information related to the topic in a coherent fashion. In all but the shortest texts, the topic will be expounded upon through the discussion of multiple subtopics. Whether the organization of the text is hierarchical in nature, as described in (Grosz and Sidner, 1986), or linear, as examined in (Skorochod'ko, 1972), boundaries between subtopics will generally exist.

In some cases, these boundaries will be explicit and will correspond to paragraphs, or in longer texts, sections or chapters. They can also be implicit. Newspaper articles often contain paragraph demarcations, but less frequently contain section markings, even though lengthy articles often address the main topic by discussing subtopics in separate paragraphs or regions of the article.

Topic boundaries are useful for several different tasks. Hearst and Plaunt (1993) demonstrated their usefulness for information retrieval by showing that segmenting documents and indexing the resulting subdocuments improves accuracy on an information retrieval task. Youmans (1991) showed that his text segmentation algorithm could be used to manually find scene boundaries in works of literature. Morris and Hirst (1991) attempted to confirm the theories of discourse structure outlined in (Grosz and Sidner, 1986) using information from a thesaurus. In addition, Kozima (1993) speculated that segmenting text along topic boundaries may be useful for anaphora resolution and text summarization.

This paper is about an automatic method of finding discourse boundaries based on the repetition of lexical items. Halliday and Hasan (1976) and others have claimed that the repetition of lexical items, and in particular content-carrying lexical items, provides coherence to a text. This observation has been used implicitly in several of the techniques described above, but the method presented here depends exclusively on it.

## Methodology

Church (1993) describes a graphical method, called *dotplotting*, for aligning bilingual corpora. This method has been adapted here for finding discourse boundaries. The dotplot used for discovering topic boundaries is created by enumerating the lexical items in an article and plotting points which correspond to word repetitions. For example, if a particular word appears at word positions $x$ and $y$ in a text, then the four points corresponding to the cartesian product of the set containing these two positions with itself would be plotted. That is, $(x, x)$, $(x, y)$, $(y, x)$ and $(y, y)$ would be plotted on the dotplot.

Prior to creating the dotplot, several filters are applied to the text. First, since closed-class words carry little semantic weight, they are removed by filtering based on part of speech information. Next, the remaining words are lemmatized using the morphological analysis software described in (Karp *et al.*, 1992). Finally, the lemmas are filtered to remove a small number of common words which are regarded as open-class by the part of speech tag set, but which contribute little to the meaning of the text. For example, forms of the verbs BE and HAVE are open class words, but are ubiquitous in all types of text. Once these steps have been taken, the dotplot is created in the manner described above. A sample dotplot of four concatenated *Wall Street Journal* articles is shown in figure 1. The real boundaries

*The author would like to thank Christy Doran, Jason Eisner, Al Kim, Mark Liberman, Mitch Marcus, Mike Schultz and David Yarowsky for their helpful comments and acknowledge the support of DARPA grant No. N0014-85-K0018 and ARO grant No. DAAL 03-89-C0031 PRI.

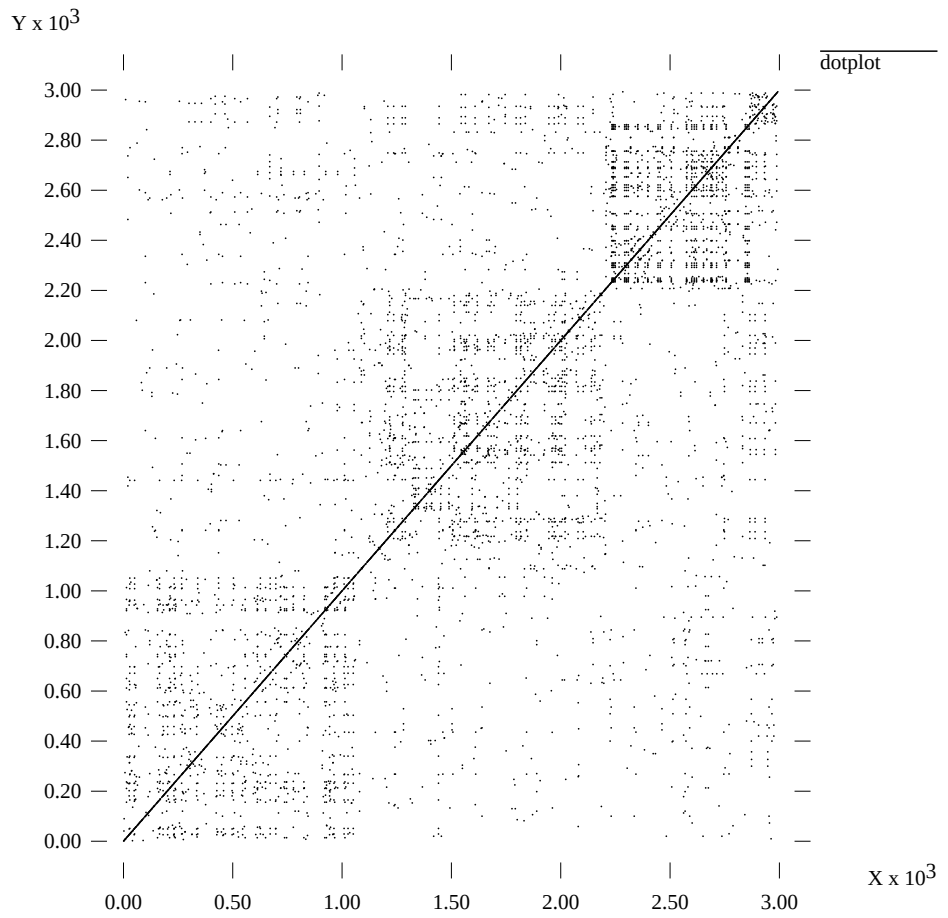

Figure 1: The dotplot of four concatenated *Wall Street Journal articles.*

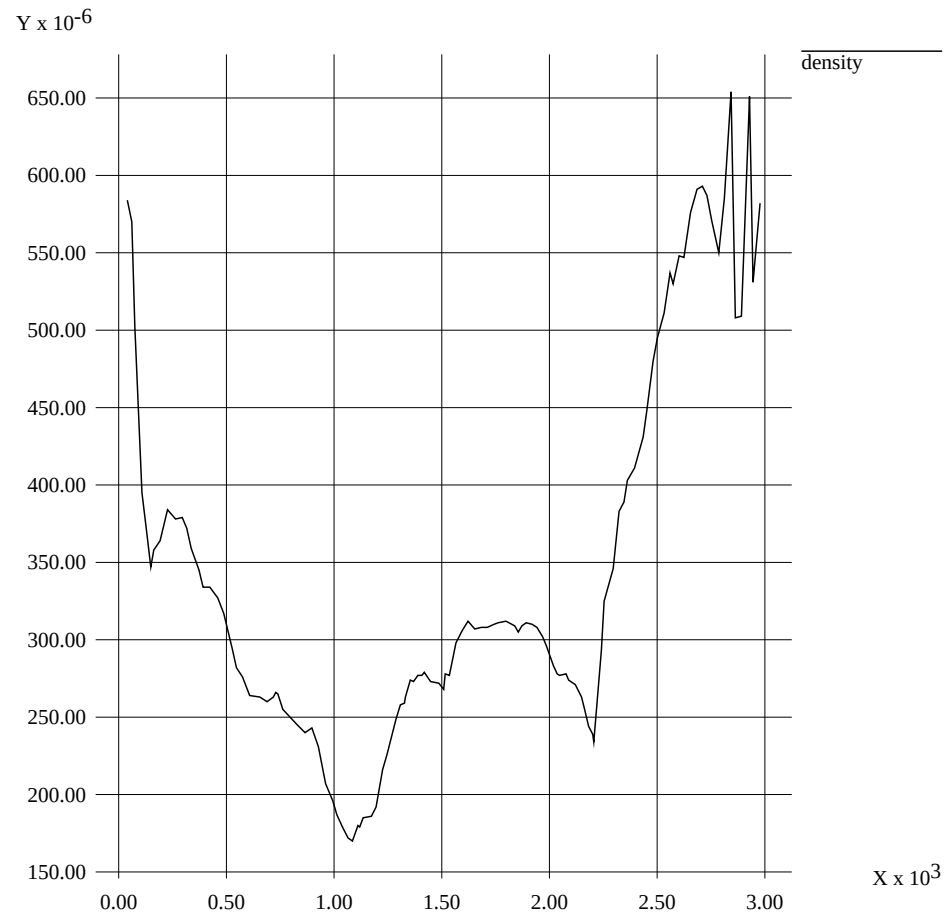

Figure 2: The outside density plot of the same four articles.

between documents are located at word positions 1085, 2206 and 2863.

The word position in the file increases as values increase along both axes of the dotplot. As a result, the diagonal with slope equal to one is present since each word in the text is identical to itself. The gaps in this line correspond to points where words have been removed by one of the filters. Since the repetition of lexical items occurs more frequently within regions of a text which are about the same topic or group of topics, the visually apparent squares along the main diagonal of the plot correspond to regions of the text. Regions are delimited by squares because of the symmetry present in the dotplot.

Although boundaries may be identified visually using the dotplot, the plot itself is unnecessary for the discovery of boundaries. The reason the regions along the diagonal are striking to the eye is that they are denser. This fact leads naturally to an algorithm based on maximizing the density of the regions within squares along the diagonal, which in turn corresponds to minimizing the density of the regions not contained within these squares. Once the densities of areas outside these regions have been computed, the algorithm begins by selecting the boundary which results in the lowest outside density. Additional boundaries are added until either the outside density increases or a particular number of boundaries have been added. Potential boundaries are selected from a list of either sentence boundaries or paragraph boundaries, depending on the experiment.

More formally, let $n$ be the length of the concatenation of articles; let $m$ be the number of unique tokens (after lemmatization and removal of words on the stop list); let $B$ be a list of boundaries, initialized to contain only the boundary corresponding to the beginning of the series of articles, 0. Maintain $B$ in ascending order. Let $i$ be a potential boundary; let $P = B \cup \{i\}$, also sorted in ascending order; let $V_{x,y}$ be a vector containing the word counts associated with word positions $x$ through $y$ in the concatenation. Now, find the $i$ such that the equation below is minimized. Repeat this minimization, inserting $i$ into $B$, until the desired number of boundaries have been located.

$$\sum_{j=2}^{|P|} \frac{V_{P_{j-1},Pj} \cdot V_{P_j,n}}{(P_j - P_{j-1})(n - P_j)}$$

The dot product in the equation reveals the similarity between this method and Heart and Plaunt's (1993) work which was done in a vector-space framework. The crucial difference lies in the global nature of this equation. Their algorithm placed boundaries by comparing neighboring regions only, while this technique compares each region with all other regions.

A graph depicting the density of the regions not enclosed in squares along the diagonal is shown in figure 2. The y-coordinate on this graph represents the density when a boundary is placed at the corresponding location on the x-axis. These data are derived from the dotplot shown in figure 1. Actual boundaries correspond to the most extreme minima—those at positions 1085, 2206 and 2863.

## Results

Since determining where topic boundaries belong is a subjective task, (Passoneau and Litman, 1993), the preliminary experiments conducted using this algorithm involved discovering boundaries between concatenated articles. All of the articles were from the *Wall Street Journal* and were tagged in conjunction with the Penn Treebank project, which is described in (Marcus *et al.*, 1993). The motivation behind this experiment is that newspaper articles are about sufficiently different topics that discerning the boundaries between them should serve as a baseline measure of the algorithm's effectiveness.

| | Expt. 1 | Expt. 2 |
|---|---|---|
| # of exact matches | 271 | 106 |
| # of close matches | 196 | 55 |
| # of extra boundaries | 1085 | 32 |
| # of missed boundaries | 43 | 349 |
| Precision | 0.175 | 0.549 |
| Precision counting close | 0.300 | **0.803** |
| Recall | 0.531 | 0.208 |
| Recall counting close | **0.916** | 0.304 |

Table 1: Results of two experiments.

The results of two experiments in which between two and eight randomly selected *Wall Street Journal* articles were concatenated are shown in table 1. Both experiments were performed on the same data set which consisted of 150 concatenations of articles containing a total of 660 articles averaging 24.5 sentences in length. The average sentence length was 24.5 words. The difference between the two experiments was that in the first experiment, boundaries were placed only at the ends of sentences, while in the second experiment, they were only placed at paragraph boundaries. Tuning the stopping criteria parameters in either method allows improvements in precision to be traded for declines in recall and vice versa. The first experiment demonstrates that high recall rates can be achieved and the second shows that high precision can also be achieved.

In these tests, a minimum separation between boundaries was imposed to prevent documents from being repeatedly subdivided around the location of one actual boundary. For the purposes of evaluation, an exact match is one in which the algorithm placed a boundary at the same position as one existed in the collection of articles. A missed boundary is one for which the algorithm found no corresponding boundary. If a boundary was not an exact match, but was within three sentences of the correct location, the result was considered a close match. Precision and recall scores were computed both including and excluding the number of close matches. The precision and recall scores including close matches reflect the admission of only one close match per actual boundary. It should be noted that some of the extra boundaries found may correspond to actual shifts in topic and may not be superfluous.

## Future Work

The current implementation of the algorithm relies on part of speech information to detect closed class words and to find sentence boundaries. However, a larger common word list and a sentence boundary recognition algorithm could be employed to obviate the need for tags. Then the method could be easily applied to large amounts of text. Also, since the task of segmenting concatenated documents is quite artificial, the approach should be applied to finding topic boundaries. To this end, the algorithm's output should be compared to the segmentations produced by human judges and the section divisions authors insert into some forms of writing, such as technical writing. Additionally, the segment information produced by the algorithm should be used in an information retrieval task as was done in (Hearst and Plaunt, 1993). Lastly, since this paper only examined flat segmentations, work needs to be done to see whether useful hierarchical segmentations can be produced.